\documentclass[letterpaper, 10 pt, conference]{ieeeconf}  

\IEEEoverridecommandlockouts                              

\usepackage{amsmath} 
\usepackage{amsfonts}
\usepackage{supertabular, booktabs, multirow, multicol, makecell}
\usepackage{subcaption}
\usepackage{graphicx}
\usepackage{pifont}
\usepackage{arydshln}

\newcommand{\cmark}{\ding{51}}%
\newcommand{\xmark}{\ding{55}}%
\newcounter{ctr}

\newenvironment{small_ind_enumerate}{\begin{list}{\thectr.}
{\usecounter{ctr}
\setlength{\rightmargin}{\rightmargin}
\setlength{\leftmargin}{0em}
\setlength{\itemsep}{\itemsep}
\setlength{\itemsep}{-0.1em}
\setlength{\topsep}{\topsep}
\setlength{\itemindent}{0.5em}
\setlength{\parsep}{\parsep}}}{\end{list}}

\title{\LARGE \bf
Explainable Knowledge Graph Embedding: Inference Reconciliation for Knowledge Inferences Supporting Robot Actions
}

\author{Angel Daruna$^{1}$, Devleena Das$^{1}$, and Sonia Chernova$^{1}$
\thanks{$^{1}$Georgia Institute of Technology, Atlanta, GA. Email: {\tt\small \{adaruna3, ddas41, chernova\}@gatech.edu}}
}

\begin{document}

\maketitle
\thispagestyle{empty}
\pagestyle{empty}

\begin{abstract}

Learned knowledge graph representations supporting robots contain a wealth of domain knowledge that drives robot behavior. However, there does not exist an inference reconciliation framework that expresses how a knowledge graph representation affects a robot's sequential decision making. We use a pedagogical approach to explain the inferences of a learned, black-box knowledge graph representation, a knowledge graph embedding. Our interpretable model, uses a decision tree classifier to locally approximate the predictions of the black-box model, and provides natural language explanations interpretable by non-experts. Results from our algorithmic evaluation affirm our model design choices, and the results of our user studies with non-experts support the need for the proposed inference reconciliation framework. Critically, results from our simulated robot evaluation indicate that our explanations enable non-experts to correct erratic robot behaviors due to nonsensical beliefs within the black-box.

\end{abstract}

\section{INTRODUCTION} \label{sec:intro}

Prior work has shown that
complex knowledge inferences afforded by learned knowledge graph representations can be used to improve a robot's robustness in ambiguous or unforeseen scenarios. Some examples include tool substitution \cite{boteanu2016leveraging} and interpolating ambiguous end-user commands \cite{nyga2018grounding}. However, these learned knowledge graph representations are usually black-boxes that are not interpretable to a non-expert user, who would require an explanation when the robot has erratic behavior due to an incorrect knowledge inference.

Explainable AI Planning (XAIP) seeks to explain an AI's reasoning to humans in sequential decision-making procedures to promote collaboration. In XAIP, inference reconciliation through dialogue with the AI is one method of explaining an AI's reasoning to a user \cite{chakraborti2020emerging}. The growing variety of questions users may ask an AI addressed by prior work include ``Why is action $a$ in plan $\pi$?", ``Why not this other plan $\pi'$?", ``Why is this policy (action) optimal?", and others. Additionally, prior work has proposed inference reconciliation frameworks that explain how plans, policies, rationales, and scene-graphs can be leveraged to explain an AI's decision making. However, to the best of our knowledge, no existing inference reconciliation framework explains how a \textit{knowledge graph representation} affects a robot's decision making.


When interacting with a user, a robot may need to justify its action based on semantic \textit{knowledge inferences} independent of the robot's plan, policy, or the scene. For example, after asking the robot to fetch their coffee, the user might ask the robot ``Why are you looking in the refrigerator?", to which the robot may reply ``Food is stored in the refrigerator, and coffee is a food”. Such an explanation not only elucidates the robot’s reasoning, but also provides a valuable opportunity for the user to correct the robot’s knowledge (e.g., ``coffee is stored in the pantry”). The aim of our work is to develop such explanation capabilities, and ultimately improve the robot's reasoning, by introducing a novel type of inference reconciliation of the form, ``Why is knowledge inference $i$, supporting action $a$, true?".

\begin{figure}
    \centering
    \includegraphics[width=\columnwidth]{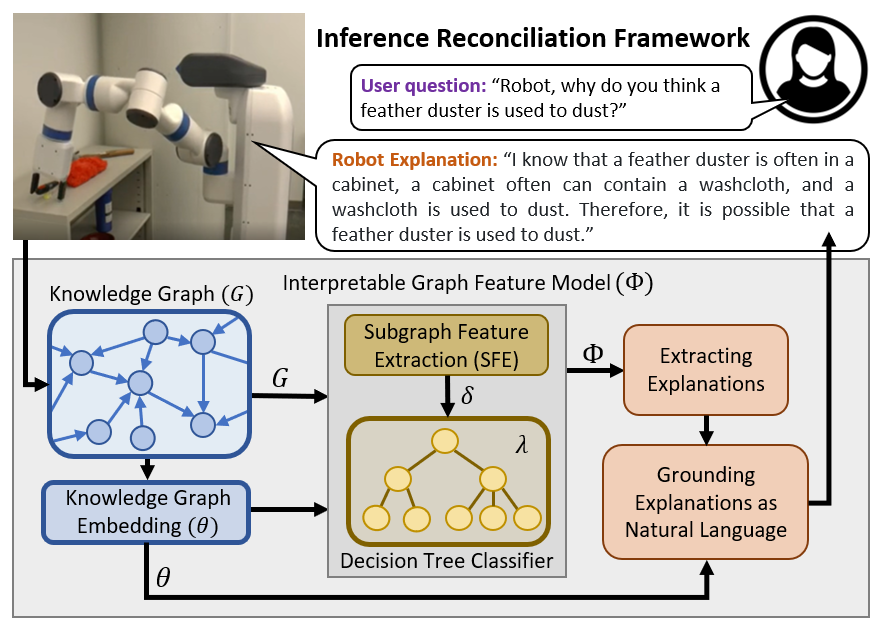}
    \vspace{-0.6cm}
    \caption{\small Overview of our inference reconciliation framework which introduces a novel explainable knowledge graph embedding method (graph feature model), leveraging decision trees, to provide natural language explanations to users.}
    \label{fig:overview}
    \vspace{-0.6cm}
\end{figure}

We introduce an inference reconciliation framework that answers a user's questions about the knowledge inference that supports a robot's action, based on knowledge graphs, XAI, and natural language (Figure~\ref{fig:overview}). Our framework uses a pedagogical XAI approach to provide explanations to non-experts about the inferences made by a learned knowledge graph representation. We develop a graph feature model as the interpretable model, using subgraph feature extraction and decision trees. We train the graph feature model to locally approximate the predictions of the learned knowledge graph representation and provide a grounded, natural language explanation for each prediction.

We evaluate our framework across three dimensions: algorithmic performance, user preference of explanations, and robot task performance. In our algorithmic evaluation, we observe statistically significant improvements in classification fidelity over baseline interpretable graph feature models, substantiating our design choices. In our user preference evaluation, we observe that users prefer our explanations as robot responses with statistically significant differences in 73.3\% of analyzed interactions. These differences in preferences validate the need for our inference reconciliation framework as it answers ``why questions" that are qualitatively different from prior work. Most importantly, in our robot task evaluation, we observe that non-expert feedback prompted by our explanations can effectively improve robot task performance (117\% and 33.7\% relative improvement in link-prediction and task execution success rate, respectively). The novelties of our work include the inference reconciliation framework as a system, our interpretable graph feature model, and our framework's ability to generate explanations that help non-experts improve robot task performance. Specifically:

\begin{small_ind_enumerate}
    \item We introduce an inference reconciliation framework that answers a novel type of user questions of the form ``Why is knowledge inference $i$, supporting action $a$, true?" using knowledge graphs;
    \item We develop and evaluate a novel graph feature model that outperforms prior work by statistically significant margins on a household knowledge dataset;
    \item We showcase a novel application of explanations within XAIP: improving downstream task performance, namely, robot behavior.
\end{small_ind_enumerate}
\vspace{-0.3cm}

\section{RELATED WORKS} \label{sec:relateds}


Our work is motivated by prior research in three areas: Knowledge Graphs in Robotics, Explainable AI Planning, and Knowledge Base Completion.

\textbf{Knowledge Graphs (KG)} are a method to model the properties and interrelations of world entities \cite{paulheim2017knowledge}. Prior works that apply KGs to robotics have demonstrated improved robustness in robot behavior by enabling robots to make complex knowledge inferences. Examples include, substituting failed demonstration actions in plans for executable actions \cite{daruna2021towards}, finding objects in alternate locations \cite{tenorth2010knowrob}, using alternative tools for tasks \cite{boteanu2016leveraging}, inferring conditional object properties \cite{LiuRSS2021}, and interpolating ambiguous end-user commands \cite{nyga2018grounding}. Most of the efforts in modeling KGs for robotics have been focused on developing computational frameworks capable of complex knowledge inferences (e.g., learning KG structure for fact prediction). Such methods have been extensively demonstrated for tasks in which robots are interacting with non-expert users (e.g., households). These two points together motivate our work, which seeks to make a robot's knowledge inferences understandable by non-experts through explanations, such that non-experts can remedy explanations of incorrect knowledge inferences, improving a robot's overall decision-making.
 
\textbf{Explainable AI Planning (XAIP)} is a focus area of Explainable AI (XAI), with the goal of explaining an AI's reasoning to humans in complex decision-making procedures to foster trust, long-term interaction, and collaboration \cite{chakraborti2020emerging}. Inference reconciliation through dialogue with the AI is one method of explaining an AI's reasoning, motivated by the notion that users have less computational power than sequential decision making systems (e.g., planners). In inference reconciliation through dialogue, user questions about the AI's decision making are answered using explanations \cite{chakraborti2020emerging}.

There are a growing variety of questions user's might ask about an AI's planning and representations affecting an AI's sequential decision making that need to be mapped into explanations as question responses. In \cite{seegebarth2012making,canal2021task}, causal link chains formed by action pre- and post-conditions within plans are used to answer ``Why is action $a$ in plan $\pi$?". In \cite{krarup2019model}, unmet properties of alternative plans (e.g., constraints) are highlighted to answer ``Why not this other plan $\pi`$?". In \cite{khan2009minimal}, the frequencies with which the current action lead to high-value future states or actions are used to answer ``Why is this policy (action) optimal?". In \cite{ehsan2019automated}, a mapping between user ascribed rationales and world states are used to answer ``Why is action $a$ taken in world state $s$?". In \cite{gobelbecker2010coming}, the transformation (excuse) to make $\Pi$ solvable is used to answer ``Why is the sequential decision making problem $\Pi$ not solvable?". In \cite{das2021semantic}, relevant scene-graph semantic relationships causing plan failure are verbalized to answer ``Why did execution of plan $\pi$ fail?".
To the best of our knowledge, no prior work in XAIP has leveraged KGs as part of the sequential decision making representations that need to be mapped into explanations as question responses. 

\textbf{Knowledge Base (Graph) Completion (KBC)} seeks to infer missing facts from a knowledge graph using existing facts \cite{nickel2015review}. In \cite{nickel2015review} two branches of KBC techniques that have received much attention are surveyed: latent and graph feature models. Latent feature models infer missing facts based on latent features of graph nodes (i.e., embeddings) and graph feature models infer missing facts based on features extracted from observed graph edges (e.g., paths). Latent feature models tend to outperform graph feature models \cite{ruschel2019explaining}. However, inferences from latent feature models are not interpretable because all embedding values are learned relative to one another and, therefore, dimensions of latent features have no inherent meaning \cite{ruschel2019explaining}. However, there exist applications where it is desirable to have accurate inferences that are interpretable (e.g., product recommendation).

Prior works have focused on improving the interpretability of embeddings for latent feature models and explainability of inferences. In \cite{xie2017interpretable,kazemi2018simple} embedding interpretability for expert users are provided through attention or importance weights over node features with respect to relation features, respectively. In \cite{yang2015embedding,zhang2019interaction}, the reasoning behind or reliability of inferences was explained to non-experts in terms of the observed short alternative paths or ``crossover interactions" between inferred and given facts, respectively. In \cite{ruschel2019explaining}, the reasoning supporting inferences was explained by learning the most highly correlated alternative paths. In \cite{nandwani2020oxkbc}, fully grounded explanations about inferences were provided by using expert labels to learn in a semi-supervised manner which ``template" explanation (similar to alternative paths) is best suited to explain an inference. We develop a novel graph feature model and compare our approach to each of these prior works in our experiments.




\section{METHODOLOGY} \label{sec:approach}


The problem of inference reconciliation is grounded in the notion that users typically have less computational ability compared to AI systems, making it difficult for users to understand an AI's solution. One solution entails providing explanations that aid in the user's inferential capabilities \cite{chakraborti2020emerging}. In our work, we consider robotic frameworks in which robot behavior is driven by knowledge inference. In particular, we assume the robot is making decisions based at least in part on a learned knowledge graph representation. To aid user understanding of robot behavior, we introduce an inference reconciliation framework \footnote{Supplementary materials: https://github.com/adaruna3/explainable-kge} that answers user questions of the form: ``Why is knowledge inference $i$, supporting action $a$, true?" (e.g. user asks a robot commanded to find a sponge, ``Why do you think you will find a sponge in the sink?"). We leverage both knowledge graphs (KG) and interpretability techniques from XAI to provide non-expert users with natural langauage explanations about a robot's knowledge inference supporting an action. As shown in Figure 1, we first gather facts about a robot's task domain, forming a KG, $\mathcal{G}$.  We use the KG $\mathcal{G}$ to learn a knowledge graph embedding (KGE), $\Theta$, that enables the robot to make complex knowledge inferences. However KGEs are black-box, and lack a mechanism to explain inferences to non-expert users (Section~\ref{sec:approach_teacher}). To make KGEs more interpretable to non-experts, we follow a pedagogical XAI approach \cite{ribeiro2016should} to provide explanations about $\Theta$'s inferences, using a graph feature model as the interpretable model, $\Phi$. Both $\mathcal{G}$ and $\Theta$ serve as inputs to $\Phi$, which performs subgraph feature extraction (SFE) and trains a decision tree classifier to locally approximate the predictions of $\Theta$ (Section~\ref{sec:approach_student}). Given a prediction from $\Theta$, we then use $\Phi$ and $\Theta$ to extract and  ground the explanation in natural language (Section~\ref{sec:approach_explain}).

\subsection{Knowledge Graph Representation} \label{sec:approach_teacher}


Knowledge graphs (KGs) are modeled as a graph $\mathcal{G}$ composed of individual facts or triples $(h,r,t)$; $h$ and $t$ are the head and tail entities (respectively) for which the relation $r$ holds, e.g., {$($\textit{cup, hasAction, fill}$)$}~\cite{chernovasituated,zhu2014reasoning,daruna2019robocse}. KGs that model real-world domains 
are
\textit{large}, \textit{sparse}, and \textit{incomplete}.  For example, a KG representing a household, while large, only represents a subset of true facts, which are sparse in a space of many potential facts. We adopt knowledge graph embeddings (KGE) for our knowledge graph representation because KGEs are designed for knowledge graphs that are large-scale and sparse \cite{nickel2015review}. Additionally, KGEs excel at learning the underlying structure of graphs to infer new facts beyond known facts in a graph (i.e. latent feature models in Section~\ref{sec:relateds}). We build upon the framework in~\cite{daruna2019robocse}, which uses a KGE to represent $\mathcal{G}$.

KGEs are distributed representations that model $\mathcal{G}$ in vector space~\cite{nickel2015review}, learning a continuous vector representation from a dataset of triples $\mathcal{D}\!=\!\big\{(h,r,t)_i,y_i|\,h_i,t_i\!\in\!\mathcal{E},r_i\!\in\!\mathcal{R},y_i\!\in\!\{0,1\}\big\}$, with $i\!\in\!\{1...|\mathcal{D}|\}$. Here $y_i$ denotes whether relation $r_i \in \mathcal{R}$ holds between $h_{i}, t_{i} \in \mathcal{E}$. Each entity $e\!\in\!\mathcal{E}$ is encoded as a vector $\textbf{v}_{e}\!\in\!\mathbb{R}^{d_{\mathcal{E}}}$, and each relation $r\!\in\!\mathcal{R}$ is encoded as a mapping between vectors $\textbf{W}_{r}\!\in\!\mathbb{R}^{d_{\mathcal{R}}}$, where $d_{\mathcal{E}}$ and $d_{\mathcal{R}}$ are the dimensions of vectors and mappings respectively~\cite{nickel2015review}. The embeddings for $\mathcal{E}$ and $\mathcal{R}$ are typically learned using a scoring function $f(h,r,t)$ that assigns higher (lower) values to positive (negative) triples~\cite{nickel2015review}. The learning objective is thus to find a set of embeddings $\Theta = \big\{\{\textbf{v}_{e}|\,e\in\mathcal{E}\},\{\textbf{W}_{r}|\,r\in\mathcal{R}\}\big\}$ that minimizes the loss $\mathcal{L}_{\mathcal{D}}$ over $\mathcal{D}$. Loss $\mathcal{L}_{\mathcal{D}}$ can take many forms depending on the KGE representation used, e.g., Negative Log-Likelihood Loss~\cite{balavzevic2019tucker}.

We make inferences (i.e. fact predictions) in KGEs by completing a transformation in the embedding space. For example, to infer tails $\{t_j|\,t_j\!\in\!\mathcal{E}\,\forall j\}$ that might complete $(h,r,\rule{3mm}{0.15mm})$, the scores $f(h,r,t_j)$ of all $j$ triples are computed, and triples with scores meeting some classification threshold are classified true. Each score $f(h,r,t_j)$ is the resultant of a sequence of high-dimensional geometric transformations between the head entity vector $\{\textbf{v}_{h}|\,h\in\mathcal{E}\}$, relation mapping $\{\textbf{W}_{r}|\,r\in\mathcal{R}\}$, and tail entity vectors $\{\textbf{v}_{t_j}|\,t\in\mathcal{E}\}$. Given the complex and relative nature of KGEs, inferences are not inherently interpretable, as discussed in Section~\ref{sec:relateds} under KBC. With the ultimate objective of providing transparent explanations of KGE inferences to non-experts, we leverage explainability techniques from XAI to explain an inference.

\subsection{Interpretable Model} \label{sec:approach_student}

Our interpretable model $\Phi$ locally approximates the inferences (i.e. fact predictions) of the KGE, denoted as $\Theta$ such that $\Phi$ can provide explanations of $\Theta$ predictions. As discussed in Section~\ref{sec:relateds}, graph feature models use graph features to infer missing facts. We develop a novel interpretable graph feature model $\Phi$ that consists of two components: interpretable features $\delta$ derived from $\mathcal{G}$ and $\Theta$, and an interpretable classifier $\lambda$ trained on $\delta$ to approximate the predictions of $\Theta$. We begin by extracting the features $\delta$ derived from a knowledge graph $\mathcal{G}$ as in \cite{gardner2015efficient}, which is formed from a dataset of triples $\mathcal{D}\!=\!\big\{(h,r,t)_i,y_i|\,h_i,t_i\!\in\!\mathcal{E},r_i\!\in\!\mathcal{R},y_i\!\in\!\{0,1\}\big\}$, with $i\!\in\!\{1...|\mathcal{D}|\}$. Here, $y_i$ denotes whether relation $r_i \in \mathcal{R}$ holds between $h_i, t_i \in \mathcal{E}$. We then train the interpretable classifier $\lambda$ over the most relevant subsets of these features to infer a triple (i.e. fact) missing from $\mathcal{G}$.

\subsubsection{Interpretable Knowledge Graph Features} \label{sec:approach_sfe}
We use Subgraph Feature Extraction (SFE) from \cite{gardner2015efficient} to extract interpretable features, $\delta$, from a graph $\mathcal{G}^\prime$, which represents a set facts believed true the the KGE $\Theta$. We begin by forming $\mathcal{G}^\prime$ using the facts in $\mathcal{G}$ classified as true by $\Theta$. We also add to $\mathcal{G}^\prime$ any facts classified as true by $\Theta$, which we form by switching the head $h$ or tail $t$ for a fact $(h,r,t)$ in $\mathcal{G}$ with the top $K$ nearest neighbor entities. Neighbors are determined by cosine similarity in the embedding space. We then used SFE to extract our interpretable features, $\delta$, from $\mathcal{G}^\prime$ instead of $\mathcal{G}$ to get a larger set of features classified as true by $\Theta$. SFE uses bi-directional breadth-first search to find all unique relation paths connecting a pair of entities in $\mathcal{G}^\prime$. Relation paths are formed from the sequence of relations that are traversed when following a path in $\mathcal{G}^\prime$ from a head entity $h$ to a tail entity $t$, where $h, t \in \mathcal{E}$. Therefore, for a relation path $\mathcal{\hat{P}}$ composed of $L$ relations, $\mathcal{\hat{P}}^{\ell}$ represents each relation on a path where $\ell\!\in\!\{1...L\}$. We encode the unique relation paths connecting entities $h,t$ paired by a relation $r$ as features using one-hot encoding.

\subsubsection{Explainable Model Training} \label{sec:approach_decision_tree}
We train a separate interpretable model $\lambda$, a decision tree, for each knowledge inference on a subset of available features that maximize $\lambda$'s classification fidelity to $\Theta$. The joint contributions of relation path combinations extracted from $\mathcal{G}^\prime$ may not be modeled as linear combinations of individual paths. 
Therefore, we use a decision tree for our interpretable model, $\lambda$, given that decision trees are able to model nonlinear decision boundaries, while remaining interpretable. Additionally, decision trees have more explicit semantics about which relation paths (i.e. features) contribute to a classification (i.e. features along the decision path), excluding extraneous relation paths that may be correlated with a class but are unnecessary to make the classification. The set of features (i.e. relation paths) selected to train a decision tree for each inference are localized to those that maximize the classification fidelity between the decision tree $\lambda$ and embedding $\Theta$. Given a fact $(h,r,t)$ to be inferred, in which $h,t\!\in\!\mathcal{E}$ and $r\!\in\!\mathcal{R}$, we implemented the locality by selecting the K nearest neighbor facts $(h_k,r,t_k)$, where $k\!\in\!\{1...K\}$, sharing a common relation $r$ such that the classification fidelity between $\Phi$ and $\Theta$ is maximized. Neighbors are determined by cosine similarity in the embedding space (e.g., $\textrm{cosine}(\textbf{v}_{h},\textbf{v}_{h_k})$).

\subsection{Extracting and Grounding Explanations} \label{sec:approach_explain}

Given a knowledge inference extracted from a user's ``why question" during inference reconciliation, we use the classifier $\lambda$ in $\Phi$ to extract the relevant relation paths, the embedding $\Theta$ to ground these relation paths to paths in $\mathcal{G}^\prime$, and templates to convert the grounded paths into natural language explanations. The knowledge inference takes the form $(h,r,t)$, where $h,t\!\in\!\mathcal{E}$ and $r\!\in\!\mathcal{R}$. Assuming $\Theta$ and $\Phi$ are in agreement about the classification of the query knowledge inference, we extract $N$ relevant relation paths $\mathcal{\hat{P}}_{n}$, where $n \in \{1...N\}$, between $h$ and $t$ that were present on the decision path when the decision tree (i.e. $\lambda$) performed classification. For each path $\mathcal{\hat{P}}_{n}$, we perform bi-directional breadth first search between $h$ and $t$ using relations from $\mathcal{\hat{P}}_{n}$ in order to ground the relation paths using $\Theta$. Therefore, for a relation path $\mathcal{\hat{P}}_{n}$ composed of $L$ relations $\mathcal{\hat{P}}^{\ell}_{n}$, where $\ell\!\in\!\{1...L\}$, the breadth first search at $h$ begins with the relation $\mathcal{\hat{P}}^{0}_{n}$ while the backward breath first search at $t$ begins with the relation $\mathcal{\hat{P}}^{L-1}_{n}$. Both recursive searches repeatedly perform inference using $\Theta$ by classifying the tails (heads) that complete the previous head (tail) and relation in $\mathcal{\hat{P}}_{n}$ for the current search step. When there is overlap between the two searches, an inference is performed that simultaneously classifies the connecting entity serving as a head and tail using $\Theta$ to ensure the connecting entity exists for the path. After completing the search, all grounded paths $\mathcal{P}_{n}$ made up of relationships classified as true by the $\Theta$ have been found and ranked in order of belief according to $\Theta$, which can be accumulated during the search. These grounded paths can then be post-processed using templates for each relation type $r\!\in\!\mathcal{R}$ and automatically corrected for grammar to produce natural-language explanations to users \cite{deemter2005real}.

\section{EXPERIMENTAL EVALUATIONS} \label{sec:experiments}


We evaluated our inference reconciliation framework with respect to algorithmic performance, user preference, and robot task performance. In our algorithmic evaluation, we used a household robot dataset to compare our interpretable graph feature model $\Phi$ with prior work and performed an ablation study on the crucial components of our algorithm design. In our user preference evaluation, we measured whether there were significant differences in user's preferences towards our explanations when a robot is asked ``why questions" during task execution. In our robot task evaluation, we measured how non-expert feedback to the robot elicited by our explanations affected robot task performance.

\subsection{Evaluation of Interpretable Graph Feature Model} \label{sec:experiments_algorithm}


Our first evaluation qualitatively and quantitatively compared our graph feature model with baseline graph feature models. Qualitatively, we considered the different features necessary for our use case that each baseline model lacked compared with ours. Quantitatively, we measured the extent to which the classifications of each considered graph feature model $\Phi$ (i.e. ours and baselines) approximates $\Theta$'s classifications (i.e. classification fidelity). Classification fidelity is a proxy measure of whether explanations produced by $\Phi$ explain $\Theta$'s reasoning \cite{ribeiro2016should,ruschel2019explaining}. For quantitative comparisons we used an evaluation procedure proposed in \cite{ruschel2019explaining} for the test split of a dataset $\mathcal{D}\!=\!\big\{(h,r,t)_i,y_i|\,h_i,t_i\!\in\!\mathcal{E},r_i\!\in\!\mathcal{R},y_i\!\in\!\{0,1\}\big\}$, with $i\!\in\!\{1...|\mathcal{D}|\}$ (see Section~\ref{sec:experiments_algorithm_quant}). Each $y_i$ denotes whether relation $r_i \in \mathcal{R}$ holds between entities $h_i, t_i \in \mathcal{E}$. We checked for significant differences in mean classification fidelity using five-fold cross-validation over $\mathcal{D}$.

\subsubsection{Qualitative Comparison} \label{sec:experiments_algorithm_qual}

We performed a qualitative comparison between graph feature models $\Phi$ from Section~\ref{sec:relateds} to select appropriate baselines for quantitative evaluation. The summary of our qualitative comparisons between all graph feature models $\Phi$ is shown in Table~\ref{tbl:xkge_methods}. We did not include SimpleE and ITransF in the quantitative comparison because relative attention weights between relations and entities are not interpretable to non-experts. Additionally, our graph feature model was designed to be embedding agnostic, allowing robotics practitioners to use the current SoTA KGE, eliminating CrossE as a baseline. Rule-Mining (DistMult) was not considered because rule-support cannot provide explanations in cases with no positively correlated relation paths because support does not reason about negative correlations between relation paths. We excluded OxKBC because instead of KG correlations, it uses expert annotations to determine which explanation is best suited for a classification, which may not provide interpretability into the KGE's (i.e. robot's) beliefs. Thus, the only prior method we quantitatively compared against is XKE, as it met all previously mentioned considerations critical to our application.

\begin{table}
    \centering
	\caption{\centering \small Comparison of Explainable KBC Methods}
	\vspace{-0.2cm}
    \begin{tabular}{|l|ccccc|} 
        \hline
        \multirow{2}{*}{\centering Method} & \multirow{2}{*}{\centering User?} & \multirow{2}{1cm}{\centering KGE Agnostic?} & \multirow{2}{1.4cm}{\centering Negative Correlations?} & \multirow{2}{1cm}{\centering Stand- alone?} & \multirow{2}{1.3cm}{\centering F1 Fidelity $(\mu,\sigma$)} \\ 
        &&&&&\\
        \hline
        SimplE \cite{kazemi2018simple} & Expert & \xmark & \cmark & \cmark & N/A \\
        ITransF \cite{xie2017interpretable} & Expert & \xmark & \cmark & \cmark & N/A \\
        CrossE \cite{zhang2019interaction} & $\neg$Expert & \xmark & \cmark & \cmark & N/A \\
        DistMult \cite{yang2015embedding} & $\neg$Expert & \cmark & \xmark & \cmark & N/A \\
        OxKBC \cite{nandwani2020oxkbc} & $\neg$Expert & \cmark & \cmark & \xmark & N/A \\
        XKE \cite{ruschel2019explaining} & $\neg$Expert & \cmark & \cmark & \cmark & $(87.9,3.4)$ \\
        \hline
        Ours & $\neg$Expert & \cmark & \cmark & \cmark & $(98.9,0.1)$ \\
        Ours ($\forall$, DT) & $\neg$Expert & \cmark & \cmark & \cmark & $(95.2,3.0)$ \\
        Ours ($\forall$, LR) & $\neg$Expert & \cmark & \cmark & \cmark & $(87.9,3.4))$ \\
        \hline
    \end{tabular}
    \label{tbl:xkge_methods}
    \vspace{-0.3cm}
\end{table}

\begin{table}
    \setlength{\tabcolsep}{4pt} 
    \vspace{0.1cm}
	\caption{\centering \small Dataset gathered from VirtualHome to learn $\Theta$}
	\vspace{-0.2cm}
    \resizebox{\columnwidth}{!}{
        \begin{tabular}{|l|ccccc|} 
            \hline
            Relation & $|E_{head}|^{\dagger}$ & $|E_{tail}|^{\dagger}$ & $|\mathcal{D}_{Tr}|^{\dagger}$ & $|\mathcal{D}_{Va}|^{\dagger}/|\mathcal{D}_{Te}|^{\dagger}$ & $|\mathcal{D}|$ \\ 
            \hline
            HasEffect       & 31 & 16   & 25 & 3 & 31 \\ 
            InverseActionOf & 12 & 12    & 12 & 1 & 14 \\
            InverseStateOf  & 16 & 16   & 14 & 1 & 16 \\  
            LocInRoom       & 43 & 4    & 86 & 10 & 106 \\ 
            ObjCanBe        & 183 & 35  & 1,369 & 171 & 1,171 \\
            ObjInLoc        & 97 & 24  & 120 & 15 & 150 \\ 
            ObjInRoom       & 183 & 4   & 334 & 41 & 416 \\ 
            ObjOnLoc        & 170 & 33  & 292 & 36 & 364 \\ 
            ObjUsedTo       & 52 & 22   & 61 & 7 & 75 \\ 
            ObjHasState     & 183 & 20  & 1,065 & 133 & 1,331 \\ 
            OperatesOn      & 52 & 188  & 1,939 & 242 & 2,423 \\
            \hline
            \multicolumn{6}{|c|}{Example entities (291 total entities)} \\
            \hline
            \small Rooms (4) & \multicolumn{5}{c|}{\small kitchen, bedroom, bathroom, livingroom} \\
            \small Locations (43) & \multicolumn{5}{c|}{\small fridge, table, sink, garbage, bed, desk, cabinet, drawer} \\
            \small Objects (189) & \multicolumn{5}{c|}{\small chair, towel, bleach, tomato, rug, plant, fork, laptop} \\
            \small Actions (35) & \multicolumn{5}{c|}{\small wipe, open, pick up, turn off, bake, unplug, disinfect} \\
            \small States (20) & \multicolumn{5}{c|}{\small dirty, clean, on, off, cooked, broken, open, plugged in} \\
            \hline
        \end{tabular}
        \label{tbl:dataset}
    }
    \footnotesize{\hspace*{\fill} $^{\dagger}$Values for an example fold of $\mathcal{D}$}
    \vspace{-0.7cm}
\end{table}

\subsubsection{Quantitative Comparison} \label{sec:experiments_algorithm_quant}

We compared our approach with XKE \cite{ruschel2019explaining} quantitatively using an evaluation procedure from \cite{ruschel2019explaining}. We first generated the inputs held constant during evaluation, the KGE $\Theta$ and dataset $\mathcal{D}$. We gathered a household robot dataset $\mathcal{D}$ of unique triples from a household simulator, VirtualHome \cite{puig2018virtualhome}, containing train $\mathcal{D}_{Tr}$, valid $\mathcal{D}_{Va}$, and test $\mathcal{D}_{Te}$ splits (Table~\ref{tbl:dataset}). We used the TuckER~\cite{balavzevic2019tucker} KGE, a recent SoTA KGE model based on the Tucker decomposition, to represent KGE $\Theta$ learned from $\mathcal{D}$. We then generated a graph feature model $\Phi$ for our approach and XKE using the same inputs, the KGE $\Theta$ and dataset $\mathcal{D}$. We then measured the classification fidelity of each $\Phi$ to $\Theta$. We measured classification fidelity as F1-Fidelity between $\Phi$ and $\Theta$ classifications, in which the KGE's classifications served as labels \cite{ruschel2019explaining}. We checked for significant differences in mean F1-Fidelity across a five-fold cross-validation over $\mathcal{D}$ using repeated-measures ANOVA and a post-hoc Tukey’s test. Please visit the footnote in Section~\ref{sec:approach} for supplementary materials detailing the implementation of $\Theta$ and each $\Phi$, the tuning of hyper-parameters, the evaluation dataset, results, and statistical analyses that are omitted here for brevity. Our results in Table~\ref{tbl:xkge_methods} show that there is a statistically significant (p=0.001) improvement in the mean F1-Fidelity between our graph feature model and XKE's.

\begin{figure}[t]
	\centering
	\includegraphics[width=0.75\linewidth]{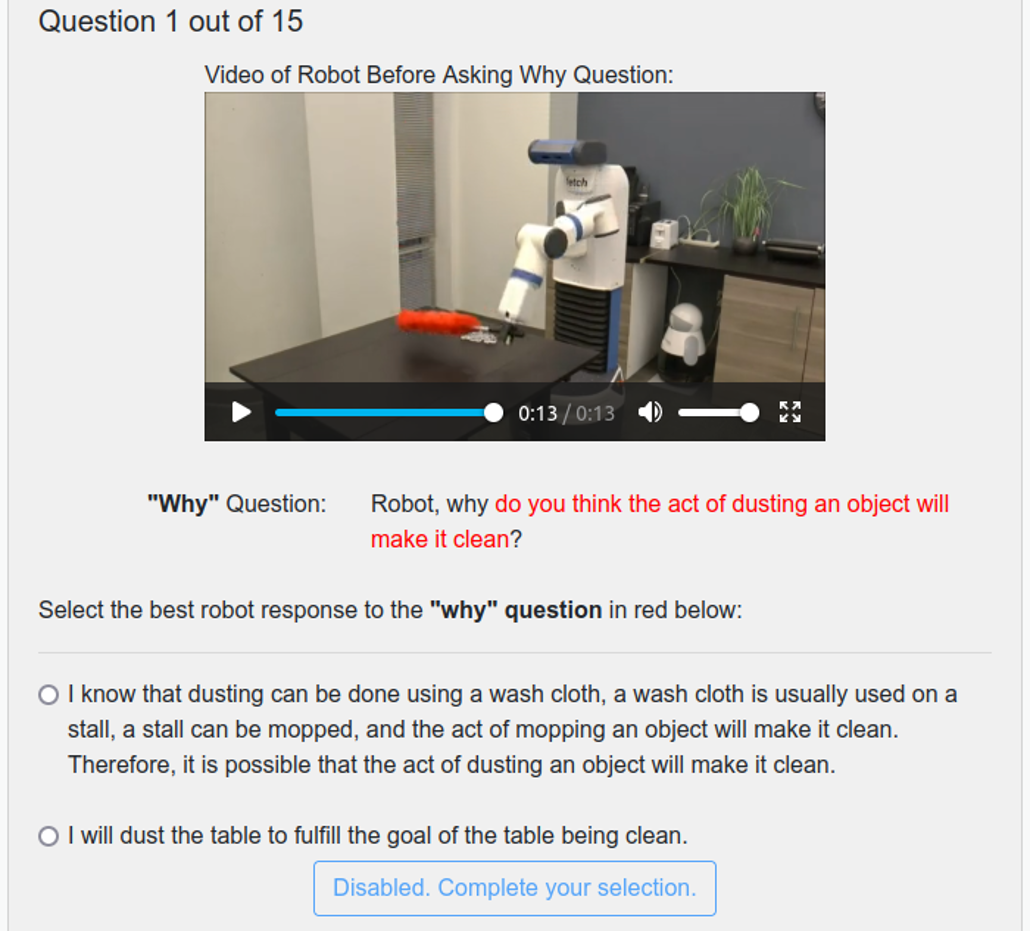}
	\vspace{-0.2cm}
	\captionsetup{width=\linewidth}
	\caption{\small Preferences User Study GUI.}
	\label{fig:preferences_gui}
	\vspace{-0.8cm}
\end{figure}

\subsubsection{Ablation Study}

We further analyzed our approach by performing an ablation study to understand how each component of our graph feature model contributed to the overall improvement in performance. We toggled two novel parts of our graph feature model not present in XKE: the use of decision trees as $\Phi$ and the locality of examples to train $\Phi$. We followed the same procedure as in the previous experiment, the results of which are in Table~\ref{tbl:xkge_methods}. The first ablation, ($\forall$, DT) in Table~\ref{tbl:xkge_methods}, shows a significant (p=0.02) drop in performance from including all available relation paths to train $\Phi$, which is still a decision tree. The fidelity of $\Phi$ drops due to the challenge of making an interpretable model approximate the global decision boundary of a black box model \cite{ribeiro2016should}. The second ablation, ($\forall$, LR) in Table~\ref{tbl:xkge_methods}, shows a significant (p=0.001) drop in performance from modeling $\Phi$ as a linear model instead of a decision tree, in addition to including all available relation paths to train $\Phi$. Here, the fidelity of $\Phi$ drops due to the non-linear joint contributions of relation paths, discussed in Section~\ref{sec:approach_decision_tree}.


\begin{table*}[t]
    \parbox{.55\linewidth}{
	    \caption{\small Preferences User Study Example Questions and Responses}
	    \vspace{-0.2cm}
        \begin{tabular}{|ll|} 
            \hline
            Knowledge Inference Why Question & Causal Why Question \\
            \hline
            \multirow{2}{5.2cm}{Why do you think you will find a washing sponge in a sink?} & Why will you move to the sink? \\
            & \\ \hdashline
            Why do you think a scrubber is used to scrub? & Why will you grab the scrubber? \\ \hdashline
            \multirow{2}{5.2cm}{Why do you think the act of scrubbing an object will make it clean?} & \multirow{2}{3.9cm}{Why will you scrub the kitchen table with the scrubber?} \\
             & \\
            \hline
            \multicolumn{2}{|c|}{Knowledge-graph-based Robot Responses} \\
            \hline
            \multicolumn{2}{|c|}{\multirow{2}{9.5cm}{I know that dusting can be done using a wash cloth, a wash cloth is usually used on a stall, a stall can be mopped, and the act of mopping an object will make it clean. Therefore, it is possible that the act of dusting an object will make it clean.}} \\ \
             & \\
             & \\
           \hdashline \multicolumn{2}{|c|}{\multirow{3}{9.5cm}{I know that a cleaning rag is often in a cabinet, a cabinet often can contain a towel, and a towel is used to wipe. Therefore, it is possible that a cleaning rag is used to wipe.}} \\
             & \\
             & \\
            \hline
            \multicolumn{2}{|c|}{Causal-link-based Robot Responses} \\
            \hline
            \multicolumn{2}{|c|}{I will sponge the table to fulfill the goal of the table being clean.} \\ \hdashline
            \multicolumn{2}{|c|}{I will move from the table to the sink to later grab the scrubber from the sink.} \\\hdashline
            \multicolumn{2}{|c|}{I will take the scrubber to later be able to scrub the table with the scrubber.} \\\hdashline
            \multicolumn{2}{|c|}{I will wipe the table to fulfill the goal of the table being clean.} \\
            \hline
        \end{tabular}
        \label{tbl:preferences_examples}
        \vspace{-0.6cm}
    }
    \quad
    \parbox{.45\linewidth}{
        \vspace{0.1cm}
        \caption{\centering \small Results of Preferences User Study}
        \vspace{-0.2cm}
        \begin{tabular}{|c|cc|l|} 
            \hline
            \multirow{2}{*}{\#} & \multirow{2}{0.6cm}{\centering Chi P-val} & \multirow{2}{0.6cm}{\centering Fisher P-val} & \multirow{2}{4.6cm}{\centering Interrupted Robot Task} \\
            & & & \\
            \hline
            1 & \textbf{0.01} & \textbf{0.006} & Find disinfectant brush on kitchen table \\
            2 & \textbf{0.01} & \textbf{0.006} & Grab disinfectant brush to disinfect \\
            3 & \textbf{0.02} & \textbf{0.01} & Grab scrubber to scrub \\
            4 & \textbf{0.00001} & \textbf{0.00001} & Find washing sponge in sink \\
            5 & 0.58 & 0.44 & Grab cleaning rag to wipe \\
            6 & 0.23 & 0.19 & Find scrubber in sink \\
            7 & \textbf{0.001} & \textbf{0.0005} & Disinfect table with disinfectant brush \\
            8 & \textbf{0.004} & \textbf{0.002} & Grab washing sponge to sponge \\
            9 & 0.13 & 0.12 & Find cleaning rag on kitchen counter \\
            10 & \textbf{0.02} & \textbf{0.01} & Grab feather duster to dust \\
            11 & \textbf{0.0001} & \textbf{0.00001} & Find feather duster in cabinet \\
            12 & 0.15 & 0.15 & Sponge table with washing sponge \\
            13 & \textbf{0.002} & \textbf{0.001} & Wipe table with cleaning rag \\
            14 & \textbf{0.009} & \textbf{0.007} & Scrub table with scrubber \\
            15 & \textbf{0.003} & \textbf{0.001} & Dust table with feather duster \\
            \hline
        \end{tabular}
        \label{tbl:preferences_results}
    }
\end{table*}

\subsection{Evaluation of Explanation Preferences} \label{sec:experiments_preferences}


Next, we evaluated our inference reconciliation framework from a non-expert's perspective. We performed a user study to characterize the relationship between different types of ``why questions" asked to a robot and a non-expert's preferred types of explanations as responses from the robot. The study evaluated two types of ``why questions" asked to discern a robot's actions during cleaning tasks: causal and knowledge inference. Our causal questions were those that inquired about the causal need for an action (e.g. ``Why will you move to the sink?").
Our knowledge inference questions were those that inquired about the underlying inferences supporting an action (e.g. ``Why do you think you will find a sponge in the sink?) (see Table \ref{tbl:preferences_examples}).
Prompted by the ``why question", users selected their preferred explanation from a list provided as possible robot responses. Our null hypothesis was that the type of ``why question" asked to a robot would not have a significant affect on the type of explanation from the robot preferred by a non-expert.

\subsubsection{Study Design} \label{sec:preferences_study_design}
We recruited 50 users from Amazon's Mechanical Turk (AMT) to perform the study, all whom were 18 years or older (M=40.7, SD=10.3). The study was between-subjects, given that each participant was randomly assigned to evaluate only one of two types of ``why questions" for each interaction. We counterbalanced for ordering effects by randomizing the question order and robot response type order. Of the 50 participants, 17 participants were filtered out for incorrectly answering any of four filtering questions (not included in results) scattered throughout the study, each which had one sensical and nonsensical response (i.e.``..take the scrubber to later slay a fire breathing dragon.."). The remaining 33 participants were used to evaluate 15 robot interactions.

In the study, users watched short videos of robot interactions within a household environment and were prompted with ``why questions" about each interaction. Specifically, in each video, the robot interacted with the world and executed one unique action towards a cleaning task before being interrupted, and the user answered a designated ``why question" that was either a causal or knowledge inference question (example provided in Figure \ref{fig:preferences_gui}). 
Users were then tasked with selecting the ``best robot response" to the ``why question". Users were provided with two different explanation types as robot responses: causal-link-based and knowledge-graph-based. The causal-link-based explanations were generated from a recent state-of-the-art plan verbalization and explanation method based on causal-link-chaining \cite{canal2021task}. The knowledge-graph-based explanations were generated by our inference reconciliation framework in Section~\ref{sec:approach} using the dataset $\mathcal{D}$ from Table~\ref{tbl:dataset}. Example questions and responses in the assessment are shown in Table~\ref{tbl:preferences_examples}. Note, we do not evaluate explanations using scene-graphs, policies, or rationales (Section~\ref{sec:relateds}), given that their assumptions do not align with our robot cleaning task).


\subsubsection{Study Results} \label{sec:preferences_study_results}
We performed a Chi-square test of independence followed by repeated Fisher's exact method measures to analyze user responses aggregated in an individual contingency table for each robot interaction in the assessment. We accounted for Type I error due to Fisher's repeated measures using Simes' alpha correction \cite{shan2017fisher}. In Table \ref{tbl:preferences_results}, we summarize the statistical analyses for each of the 15 robot interactions analyzed, excluding the filtering question instances. We observed that when asked a knowledge inference ``why question", participants select a knowledge-graph-based robot response more often than not in all questions except 5, and that 73.3\% of these selections were significant (in \textbf{bold}). Similarly, when asked causal ``why questions", participants select a causal-link-based robot response more often than not in all questions, and 73.3\% of these differences are significant. Overall, our results indicate that non-experts recognize the qualitative differences in the two types of ``why questions", which tend to significantly effect their preferred type of robot response. In other words, there is a need for knowledge-based robot responses given they are better suited for knowledge inference ``why questions".  

\subsection{Validation of Explanations for Downstream Tasks} \label{sec:experiments_feedback}


In our final experiment we evaluated whether non-experts could use our inference reconciliation framework to improve robot task performance. The robot's task was \textit{robust task execution}, wherein the robot is provided a demonstrated task-plan recorded in a demonstration environment, and asked to generalize the task-plan to new execution environments. Given that robots cannot be assumed to be error-free, we relaxed the assumption from prior experiments that the robot had gathered a high-quality dataset by including robot perception noise (i.e. the robot had a faulty object and affordance detector). We incorporated robot perception noise into a new dataset $\mathcal{\hat{D}}$ by randomly corrupting 30\% of facts for each relation in the dataset $\mathcal{D}$ from prior experiments (Table~\ref{tbl:dataset}). As a consequence, the KGE $\hat{\Theta}$ learned from the $\mathcal{\hat{D}}$ dataset often makes nonsensical knowledge inferences (e.g., $(\textrm{sponge},\textrm{ObjUsedTo},\textrm{microwave})$) that lead to erratic robot behavior when executing tasks. We validated whether nonsensical facts supporting inferences made by $\hat{\Theta}$ can be revealed to non-experts using explanations generated by our inference reconciliation framework. We hypothesized that if non-experts can accurately recognize and correct nonsensical facts in explanations, then that feedback can be used to improve $\hat{\Theta}$, and in turn, improve robot behavior.

We performed a user study to measure how well non-experts can correct nonsensical facts within natural language explanations (i.e. correction accuracy). The explanations were generated by our inference reconciliation framework. Specifically, the interpretable model $\Phi$ was trained on $\mathcal{\hat{D}}$ for incorrect classifications of facts in $\mathcal{D}_{Te}$ where $\hat{\Theta}$ and $\Phi$ provided the same classification (see Section~\ref{sec:approach}).
The results of the user study informed a confidence interval for the expected non-expert correction accuracy. 


\subsubsection{Study Design} \label{sec:experiments_feedback_study_design}
\begin{figure}[t]
	\centering
	\includegraphics[width=0.8\linewidth]{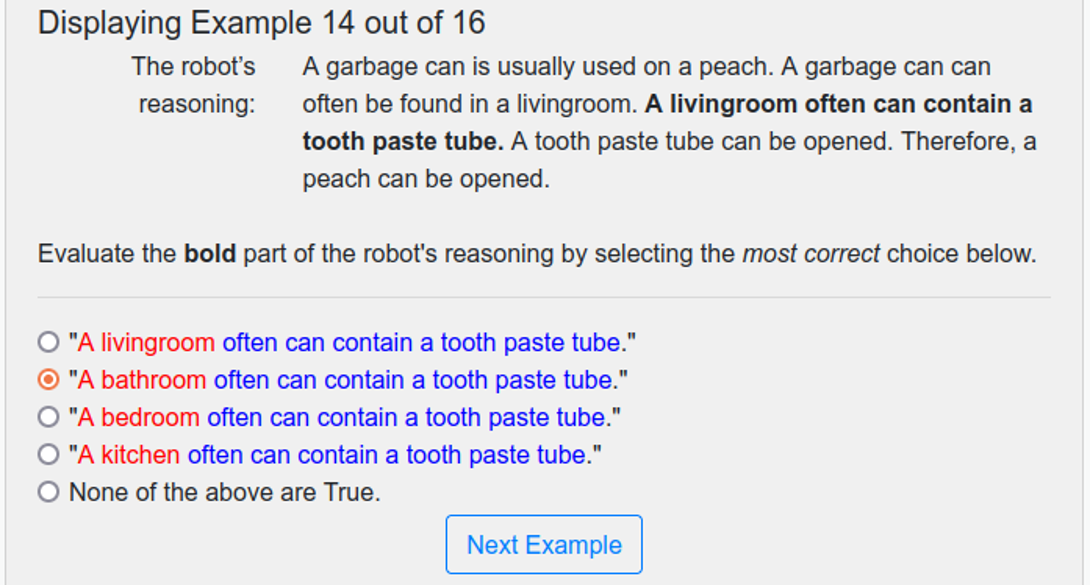}
	\vspace{-0.1cm}
	\captionsetup{width=\linewidth}
	\caption{\small Non-expert Feedback User Study GUI.}
	\label{fig:feedback_gui}
	\vspace{-0.8cm}
\end{figure}

We recruited 19 participants from AMT, all of whom were 18 years or older (M=33.4, SD=5.9). Of the 19 participants, 1 was filtered out for performing below chance on the practice portion of the study.
In the study, users were tasked with identifying and correcting nonsensical facts for a series of knowledge inferences and accompanied explanations. Specifically users traversed through each supporting fact of the explanation and selected the ``most correct" fact from a list of options, which included the supporting fact from the explanation, three other facts classified as most true by the KGE that could replace the supporting fact from the explanation, and ``None of the above" (example in Figure~\ref{fig:feedback_gui}).


We formed a confidence interval for the expected non-expert correction accuracy because exhaustively recruiting users to evaluate all grounded explanations generated for misclassified examples in 
$\mathcal{\hat{D}}_{Te}$ was impractical due to the large number of grounded explanations (10,000+). Instead, we randomly assigned each user 16 questions sampled from the large set of grounded explanations. In total, 18 users answered 96 sampled questions. We ensured that each sampled question received three responses (randomly sampled), and combined these responses using majority voting to create 6 meta-users.
We determined the necessary population size of meta-users to be 6, by forming a 95\% confidence interval with a 5\% error margin using the sample standard-deviation of meta-user correction accuracy.


\subsubsection{Study Results} \label{sec:experiments_feedback_study_results}

The mean non-expert correction accuracy was 86.6\% with an error margin of 4.1\%. We confirmed non-expert correction accuracy samples were normally distributed using Shapiro-Wilk's Test (p=0.83). Additionally, to ensure no grouping bias, we performed a one-way ANOVA to test that there were no significant differences in the mean performances of users whose combined responses formed each meta-user (p=0.72).

\subsubsection{Improvement of KGE} \label{sec:experiments_feedback_kbc_results}
We characterized the effect of non-expert correction accuracy on improving the KGE by measuring inference performance. Our initial KGE was $\hat{\Theta}$ learned from $\mathcal{\hat{D}}$. We corrected 86.6\%
of the corrupted facts in $\mathcal{\hat{D}}$ to form a dataset $\mathcal{\bar{D}}$ and learned a new KGE, $\bar{\Theta}$, from $\mathcal{\bar{D}}$. Finally, we measured the difference in inference performance between $\hat{\Theta}$ and $\bar{\Theta}$ on our original dataset $\mathcal{D}$ (from Table~\ref{tbl:dataset}) for the common KGE task, link-prediction \cite{nickel2015review}.

In short, the link-prediction evaluation task is to rank complete triples from incomplete ones in test splits, i.e., rank heads $h$ given $(r, t)$ or tails $t$ given $(h, r)$. To perform link-prediction, each test triple $(h,r,t)$ is first corrupted by replacing the head (or tail) entity with every other possible entity $\mathcal{E}$. Then all corrupted triples that represent a valid relationship between the corresponding entities are removed to avoid underestimating the embedding performance, known as ``filtered setting" \cite{nickel2015review}. Last, scores are computed for each test triplet and its (remaining) corrupted triplets using the scoring function $f(h,r,t)$, then ranked in order of belief. For each test triple $(h,r,t)$, the mean reciprocal rank of the test triple is calculated as a measure of inference performance.

\begin{figure}[t]
	\centering
	\includegraphics[width=0.93\linewidth]{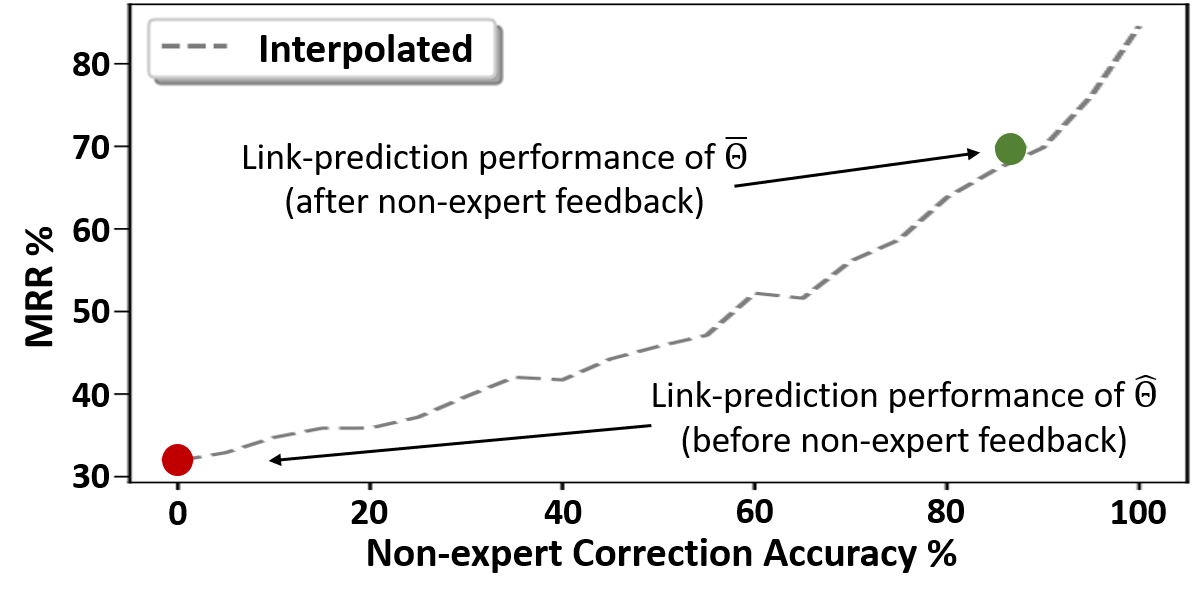}
	\vspace{-0.4cm}
	\captionsetup{width=\linewidth}
	\caption{\small Effect of Non-expert Correction Accuracy on MRR\%.}
	\label{fig:feedback_mrr}
	\vspace{-0.8cm}
\end{figure}

We observed that the mean non-expert correction accuracy provided a 117\% relative improvement in MRR between $\hat{\Theta}$ and $\bar{\Theta}$ for the link-prediction task. Shown in Figure~\ref{fig:feedback_mrr} is the MRR of $\hat{\Theta}$ as a red dot (32.1\%) and $\bar{\Theta}$ as a green dot (69.9\%). Figure~\ref{fig:feedback_mrr} also includes interpolated results for the range of possible non-expert correction accuracy.

\subsubsection{Improvement of Robot Behavior} \label{sec:experiments_feedback_robot_results}

As a final result, we characterized the effect of non-expert correction accuracy on improving the robot's behavior by measuring robot success rate. We considered a real-world scenario in which a household robot performs robust execution of household cleaning tasks, using the system and definitions in \cite{daruna2021towards}. In robust task execution, the robot is provided an execution environment $E_x$ and a demonstrated task plan $T_d$ that is recorded in a demonstration environment $E_d$, and asked to find a modified task plan $T_x$, executable in $E_x$, that accomplishes the goal(s) of $T_d$. Task plans are defined as a sequence of primitive actions, where each primitive action may or may not be parameterized by objects. Task plans are incrementally modified using knowledge inferences from a KGE with the system defined in \cite{daruna2021towards} to find an executable task plan. Execution environments are sampled by perturbing the object used to demonstrate the task; changing the object’s location, type, or both. The perturbations are made in accordance with the action-object-location distributions present in $\mathcal{D}$, ensuring that objects are not placed at implausible locations (e.g., broom inside the toilet) and that the intended generalization is not unreasonable (e.g. cleaning a table with a washing-machine). As a result, the demonstrated task plan often fails due to unsatisfied pre-conditions of primitive actions, and the robot must generalize the demonstrated task plan to formulate the \textit{executable} task plan.

We measured the improvement in robot success rate provided by the mean non-expert correction accuracy by deploying $\hat{\Theta}$ and $\bar{\Theta}$ on a simulated robot in a simulation household environment performing robust execution of cleaning tasks. We sampled 50 initial demonstrations with 10 executions environments in each, for a total of 500 robot executions and measured the robot's execution success rate when using $\hat{\Theta}$ and $\bar{\Theta}$. Our experiment showed a 33.7\% relative improvement in the robot's success rate due to the non-expert feedback (i.e. the robot succeeded in 187 sampled environments using $\hat{\Theta}$ and 250 using $\bar{\Theta}$). We measured the success rate of a robot that only repeats the default demonstration and a robot selecting random generalizations of the task plan as reference points for the difficulty of the task, which were 9 and 22 successes, respectively.

\begin{table}[t]
    \centering
    \caption{\centering \small Corrected Robot Behaviors}
    \vspace{-0.2cm}
    \begin{tabular}{|c|l|} 
        \hline
        Action & Objects robot attempted to use to perform action \\
        \hline
        dust & computer, radio \\
        wipe & crayon \\
        disinfect & toaster, television, candle, pillow  \\
        \hline
    \end{tabular}
    \label{tbl:robot_study_qual}
    \vspace{-0.6cm}
\end{table}

In addition to the quantitative improvement in success rate, we observed qualitative changes in the robot's generalization behaviors during task execution. Examples are shown in Table~\ref{tbl:robot_study_qual}. Each combination of action and object in Table~\ref{tbl:robot_study_qual} is an instance of a robot behavior that was corrected by the non-experts. For example, when using $\hat{\Theta}$, the robot attempted to dust using a radio. However, by using $\bar{\Theta}$, which incorporates the non-expert feedback, those nonsensical task generalizations no longer are attempted.

\section{CONCLUSIONS}

In summary, we introduce an inference reconciliation framework that answers a user's questions about knowledge inferences supporting robot actions using knowledge graphs, XAI, and natural language. Our framework follows a pedagogical XAI approach, by using an interpetable graph feature model to locally approximate the classifications of a black-box model, a KGE. Through a three-fold evaluation, we demonstrate the importance of our framework both with respect to interpretability as well as improved task performance. Specifically, we show via an algorithmic evaluation that leveraging a decision tree classifier as an interpretable graph feature model in our framework leads to higher F1-Fidelity compared to prior use of linear regression models for explainable KGEs. Additionally, through user evaluations, we demonstrate that our explanations are highly preferred and accessible. Through a user preference evaluation, we demonstrate a significant preference towards our framework's explanations for knowledge inference ``why questions." Additionally, when relaxing the assumption that robots are error-free, we showcase the effectiveness of our explanations in helping users identify and correct nonsensical beliefs in robots' knowledge representations, consequently improving robot task performance.

\bibliographystyle{IEEEtran}
\bibliography{IEEEabrv,references}

\end{document}